\documentclass{article}

\usepackage{arxiv}

\usepackage{cite}
\usepackage{amsmath,amssymb,amsfonts}
\usepackage{array,multirow}
\usepackage{float}
\usepackage{algorithmic}
\usepackage{graphicx}
\usepackage{textcomp}
\usepackage{longtable}

\usepackage{array,subfigure}
\usepackage{ amssymb }
\usepackage{ mathrsfs }
\usepackage{mathtools}
\usepackage{commath}
\usepackage[sc,osf]{mathpazo}
\usepackage{arydshln}

\usepackage[utf8]{inputenc} % allow utf-8 input
\usepackage[T1]{fontenc}    % use 8-bit T1 fonts
\usepackage{hyperref}       % hyperlinks
\usepackage{url}            % simple URL typesetting
\usepackage{booktabs}       % professional-quality tables
\usepackage{amsfonts}       % blackboard math symbols
\usepackage{nicefrac}       % compact symbols for 1/2, etc.
\usepackage{microtype}      % microtypography
\usepackage{lipsum}

\title{Dueling Deep Q-Network for Unsupervised Inter-frame Eye Movement Correction in Optical Coherence Tomography Volumes}
\author{Yasmeen George \thanks{Y. George, S. Sedai, B. Antony, and R. Garnavi are with IBM Research Australia  (e-mails: georgey@ibm.com, {ssedai,bhavna.antony, rahilgar}@au1.ibm.com)} \and Suman Sedai\footnotemark[1] \and Bhavna J. Antony\footnotemark[1] \and Hiroshi Ishikawa\thanks{H. Ishikawa, G. Wollstein, J. Schuman are with Department of Ophthalmology, NYU Langone Health, New York, United States (e-mails: {hiroshi.Ishikawa,gadi.wollstein,joel.schuman}@nyulangone.org)} \and Gadi Wollstein \footnotemark[2] \and Joel S. Schuman\footnotemark[2] \and  Rahil Garnavi\footnotemark[1]}

\begin{document}
\maketitle

	\begin{abstract}
		In optical coherence tomography (OCT) volumes of retina, the sequential acquisition of the individual slices makes this modality prone to motion artifacts, misalignments between adjacent slices being the most noticeable. Any distortion in OCT volumes can bias structural analysis and influence the outcome of longitudinal studies. On the other hand, presence of speckle noise that is characteristic of this imaging modality, leads to inaccuracies when traditional registration techniques are employed. Also, the lack of a well-defined ground truth makes supervised deep-learning techniques ill-posed to tackle the problem. In this paper, we tackle these issues by using deep reinforcement learning to correct inter-frame movements in an unsupervised manner. Specifically, we use dueling deep Q-network  to train an artificial agent to find the optimal policy, i.e. a sequence of actions, that best improves the alignment by maximizing the sum of reward signals. Instead of relying on the ground-truth of transformation parameters to guide the rewarding system, for the first time, we use a combination of intensity based image similarity metrics. Further, to avoid the agent bias towards speckle noise, we ensure the agent can see retinal layers as part of the interacting environment. For quantitative evaluation, we simulate the eye movement artifacts by applying 2D rigid transformations on individual B-scans. The proposed model achieves an average of 0.985 and 0.914 for normalized mutual information and correlation coefficient, respectively. We also compare our model with elastix intensity based medical image registration approach, where significant improvement is achieved by our model for both noisy and denoised volumes.
		
	\end{abstract}
	
	\section{Introduction}
	Optical coherence tomography  (OCT) technology provides clinicians with real-time and high-resolution images of ocular structures which are of great use in diagnosing and monitoring retinal diseases, evaluating progression, and assessing response to therapy \cite{fujimoto2016development}.  Three generations for OCT since it was invented in 1991 \cite{huang1991optical} and over the past few decades, different commercially available  OCT instruments were developed. Each device is characterized by several parameters such as lateral and axial resolutions, penetration depth and imaging speed. Fast image acquisition of OCT volumes, i.e. A-scan rate, is very important to reduce retinal motion. Yet, it is limited by the camera read-out rate in OCT scanner device \cite{sanchez2019review}. Another type of motion artifacts is the axial eye motion with a frequency varies from 3 to 12 Hz and its exact mechanism has not  been clearly known. In addition to the involuntary fixational eye movements that are usually categorized as high frequency tremors, rapid microsaccades, or slow drifts, depending on their frequency and magnitude. These types of motion artifacts during OCT volume scanning result in deformed 3D data of the retina \cite{spaide2015image}. The correction of distorted data is very essential to improve OCT image quality and therefore better diagnoses.
	
	Retinal motion vary on amplitude, direction, and frequency, making its combination difficult to predict \cite{sanchez2019review}. Moreover, retinal motion may differ significantly between individuals, hindering the development of a generalized theoretical or learning models for retinal motion prediction. This issue can be tackled by using advanced deep learning (DL) models with a large OCT dataset that covers different types of retinal motion. In this regard, ground-truth for inter-frame misalignment is needed, which is very expensive and difficult to manually annotate. Further, unsupervised learning models with OCT retinal cubes are greatly biased towards the speckle noise misleading the final outcomes of the alignment.
	%How we will solve it? RL - Unsupervised
	In this work, we propose an unsupervised inter-frame movement correction approach that works really well even in the presence of speckle noise.  The approach is based on deep reinforcement learning, in which an artificial agent is trained using deep q-network to find a strategy of sequential actions that best improves the alignment between 2d-slices in OCT data. 
	
	\textbf{Related Work.}
	The literature eye movement correction methods can be divided into feature-based and intensity-based approaches. Feature-based registration methods use landmarks of the image, such as vasculature, vessel intersections, and retinal layers to correct OCT data misalignment, while intensity based approaches rely on the similarity between images such as correlation and mutual information \cite{sanchez2019review,baghaie2015state,baghaie2017involuntary}. 
	Further, the literature also presents the use of deep reinforcement learning for medical image registration \cite{liao2017artificial,krebs2017robust}, where the agent uses ground truth transformation parameters for training. 
	\newline \\
	We summarize our contributions as follow:
	%\vspace{-3ex}
	\begin{itemize}
		\item For the first time, we propose a dueling deep q-network for OCT inter-frame image alignment (DDQN-OCT) that does not require landmarks or transformation parameters ground truth. 
		\item We use a combined intensity based image similarity metrics to guide the rewarding system for training the artificial agent in an unsupervised fashion. 
		\item The proposed approach does not require the removal of speckle noise, which is a common preprocessing step for all 2D and 3D OCT registration methods. 
		\item  Our approach has significant improvement over elastix intensity based medical image registration.
	\end{itemize}
	\section{Methodology}
	\subsection{Reinforcement Learning Framework}
\begin{figure}
		\centering
		\includegraphics[width=1\linewidth]{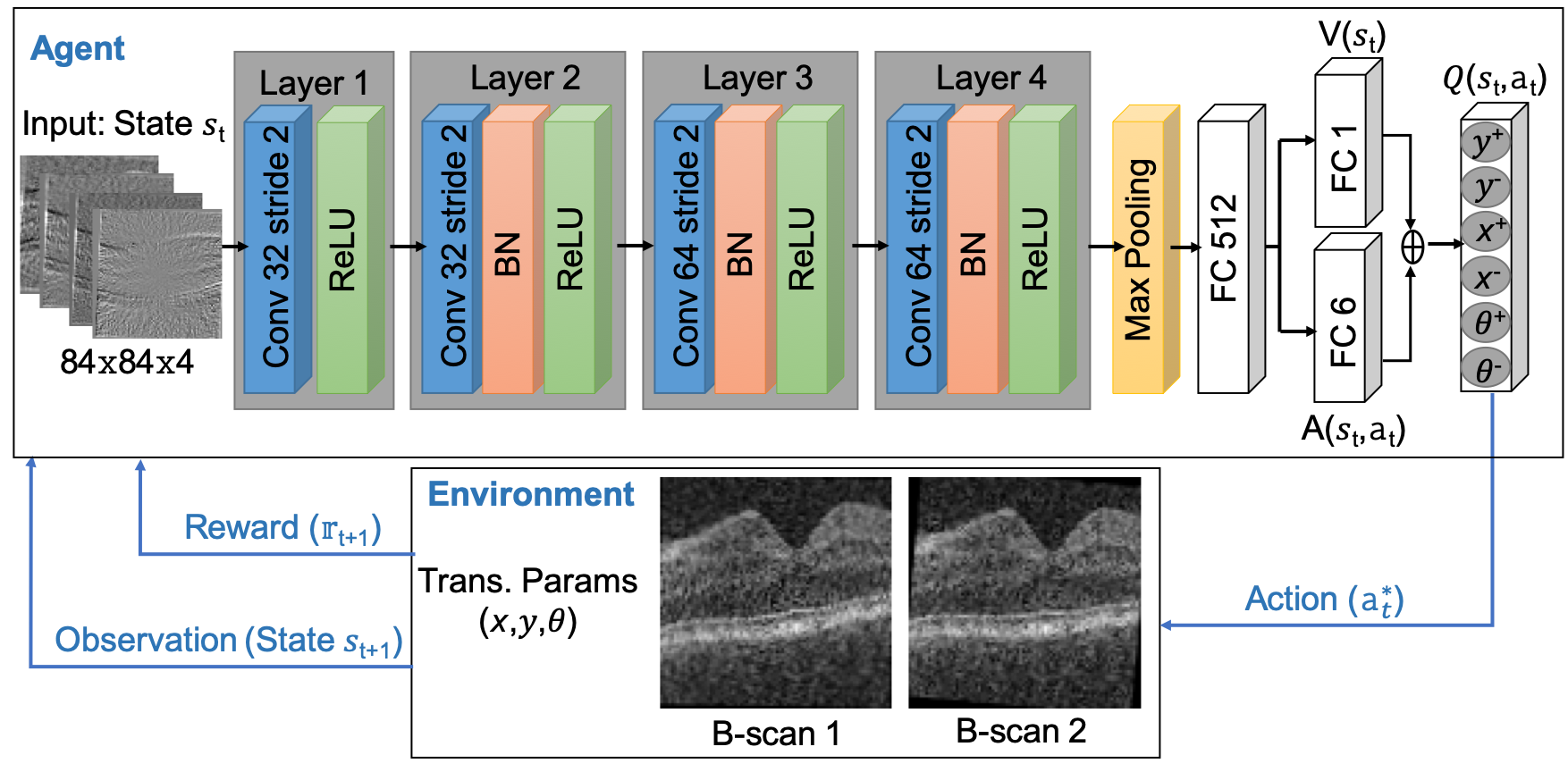}
		%\vspace{-3ex}
		\caption{Deep reinforcement learning framework for eye motion correction in OCT volumes}
		\label{fig:drlframework}
	\end{figure}

	We formulate the inter-frame movement correction in OCT volumes as a 2D rigid registration problem that matches two adjacent B-scans in the fast scanning plane  ($\mathcal{F}_{B_i}$ and $\mathcal{F}_{B_{i+1}}$). This is accomplished by finding the optimal spatial transformation $\mathcal{T}$ that aligns $\mathcal{F}_{B_{i+1}}  \circ \mathcal{T}_i$ with $\mathcal{F}_{B_i}$.  The 2D-rigid transformation $\mathcal{T} $ has 3 parameters: two translations ($\mathcal{T}_x,\mathcal{T}_y$) and one rotation ($\mathcal{T}_\theta$). For this purpose, we use deep reinforcement learning (DRL) to solve this optimization problem in an unsupervised manner. Figure \ref{fig:drlframework} shows the framework of the proposed approach. Specifically, an artificial agent learns by interacting with an environment ($E$) to maximize the cumulative reward signals (${R}$) throughout the agent's lifetime. 
	At every iteration $t$, given a state $s_t \in \mathcal{S}$ that represents the difference image of the two adjacent B-scans, the agent selects an action $a_t \in \mathcal{A}$ that is associated with a scalar reward signals ${R}_t$. $\mathcal{S}$ is the set of states that the agent can see and $\mathcal{A}$ is the set of discrete actions that the agent can take. $\mathcal{A}$ consists of 6 candidate transformations that lead to the change of $\pm 1$ in one parameter of $\mathcal{T}$. 
	During training, the agent learns the optimal policy $\Pi$ (i.e. a strategy of sequential actions) that maps a current state $s_t$ to an optimal action $a_t^*$ that best improves the alignment by maximizing the sum of reward signals seen over the agent's lifetime. The optimal action-selection policy is identified by learning an action value function $\mathcal{Q}(s,a)$ (i.e. $\mathcal{Q}$-function) that measures the quality of taking an action $a_t$ given a state $s_t$, as defined by Watkins et al. \cite{Watkins1992}. The $\mathcal{Q}$-function can be solved using Bellman iterative approach \cite{bellman2013dynamic} as in Equation \ref{eq:action-value}. 
	\begin{equation}
	\begin{split}
	\mathcal{Q}_{t}(s_t,a_t) & = E [R_{t} + \gamma R_{t+1} +...+ \gamma^{n-1} R_{t+n}| s_t,a_t] \\
	& =  E [R_{t} + \gamma \max_{a_{t+1}} Q_{t+1} (s_{t+1},a_{t+1})] 
	\end{split}
	\label{eq:action-value}
	\end{equation}
	where $\gamma \in [0,1]$ is the future rewards discount factor. $s_{t+1}$ and $a_{t+1}$ are the next state and action.
	
	\subsection{Dueling Deep Q-Network for Optimal Policy Estimation}
	In this paper, we follow Mnih et al. \cite{mnih2015human} who proposed a deep Q-network (DQN) to approximate the action-value function using deep neural network (DNN) as in Equation \ref{eq:dqn_mnih}.  
	
	\begin{equation}
	Q(s_t,a_t) =
	\begin{cases} 
	R(s_t, a_t) + \gamma Q(s_{t+1},a_{t+1}^*) & \text{if  $D(\mathcal{F}_{B_i},\mathcal{F}_{B_{i+1}}  \circ \mathcal{T}_t) > \epsilon$}\\
	R(s_t, a_t) + R & \text{Otherwise} 
	\end{cases} 
	\label{eq:dqn_mnih}
	\end{equation} 
	where $R$ is a bonus reward value the agent receives when it finds the best alignment. This is reached when the distance between the reference B-scan and the aligned B-scan is within the distance threshold $\epsilon$. The immediate reward $R(s_t, a_t)$ for a state-action pair is calculated using Equation \ref{eq:reward}.  
	\begin{equation}
	R(s_t, a_t) = \left\lVert  \mathcal{D}(\mathcal{F}_{B_i} ,\mathcal{F}_{B_{i+1}}  \circ \mathcal{T}_{t-1})   - \mathcal{D}(\mathcal{F}_{B_i} ,\mathcal{F}_{B_{i+1}}  \circ \mathcal{T}_{t})   \right\rVert 
	\label{eq:reward}
	\end{equation}
	where $\mathcal{T}_{t-1}$ and $\mathcal{T}_t$ refer to the transformation values before and after action $a_t$ is selected that is parameterized by $[t_x,t_y, \theta]$. $\mathcal{D}$ is the dissimilarity metric.
	
	Finding a good measure of dissimilarity is very crucial for the success of the agent learning process. The literature papers in this area rely on the ground truth transformation parameters \cite{liao2017artificial,krebs2017robust}, where  $\mathcal{D} = T - T_{\text{gr}}$. In our work, we propose the use of intensity based image similarity metrics to train the agent in an unsupervised manner that was not proposed in the literature before. The proposed dissimilarity metric is based on two statistical measures, namely correlation coefficient ($\rho$) and structural similarity index measure (SSIM) as in Equation \ref{eq:similarity}. $\rho$ measures the degree of change in one causes the change in the other, while SSIM considers the perceptual likeness and structural information. 
	
	\begin{equation}%\begin{split}
	\begin{gathered}
	\mathcal{D}(x,y) = 1- \frac{\rho_{xy} + \text{SSIM}(x,y)}{2} \\ 
	\rho_{xy}=  \frac{\sigma_{xy}}{\sigma_x\sigma_y} \\ 
	\text{SSIM}(x,y) = \frac{(2\mu_{x}\mu_{y}+c_1)(2\sigma_{xy}+c_2)}{(\mu_x^2+\mu_y^2+c_1)(\sigma_x^2+\sigma_y^2+c_2)} %\end{split}
	\end{gathered}
	\label{eq:similarity}
	\end{equation}
	where $\mu_{x}$ and $\sigma_x$ are the average and variance of reference image x, and $\mu_{y}$ and $\sigma_y$ are the average and variance of the transformed image y. Also, $\sigma_{xy}$ refers to the covariance between x and y. $c_1$ and $c_2$ are stabilization variables.
	
	We also adopt the action-state value function split notion, proposed by Wang et al. \cite{wang2015dueling}, called dueling DQN. In which, $Q(s,a)$ is decomposed into action-independent ($V(s)$) and action-dependent value ($A(s,a)$) functions. This has shown to provide robust state value estimates. 
	The architecture of our DNN network is shown in Figure \ref{fig:drlframework}, which consists of 4 convolutional layers, each is followed by batch normalization 
	and ReLU activation, except the first convolutional layer which does not have batch normalization. The convolutional layers have incremental number of the filters of 32-32-64-64 with kernel sizes of 5-5-4-3 in order, and stride of 2 for all layers. This is followed by 2D max-pooling layer with size of 2 and a fully-connected layer with 512 nodes. Then the output of fully-connected layer is passed to two branches for action-dependent and action-independent value functions with 6 and 1 nodes in order. Finally, a fully-connected layer connects the sum of the two branches to the output layer with 6 nodes, each corresponds to one of the actions in $A$. The input to the network is computed by subtracting the reference B-scan and the transformed one, i.e. $I = \mathcal{F}_{B_i} - \mathcal{F}_{B_{i+1}}$ for each of the previous $N$ steps. This is to obtain more stable search trajectories and prevent the agent from oscillation issue. Also, the loss function is calculated using mean squared error as shown in Equation \ref{eq:dqn}.
	\begin{equation}
	\mathcal{L}_{\text{\tiny{DQN}}} = E [(R_{t} + \gamma \max_{a_{t+1}} Q(s_{t+1},a_{t+1}) - Q(s_{t},a_{t})  )^2] 
	\label{eq:dqn}
	\end{equation}
	
	\section{Data and Implementation Details}
	\textbf{\textit{Dataset.}} The dataset contains 10,370 OCT macular scans from both eyes of 1678 individuals, acquired on a Cirrus SD-OCT Scanner (Zeiss; Dublin, CA, USA) over multiple visits. The dataset has 427 healthy scans from 109 individuals and the other scans with different ocular conditions including glaucoma, optic neuropathy, plateau iris and others. 
	The scans have 200$\times$200$\times$1024 (a-scans$\times$b-scans$\times$depth) voxels per cube covering an area of 6$\times$6$\times$2 ${mm}^3$. This is an observational study that was conducted in accordance with the tenets of the Declaration of Helsinki and the Healthy Insurance Portability and Accountability Act. %%%Blind The Institutional Review Board of New York University and the University of Pittsburgh approved the study, and all subjects gave written consent before participation. 
	\newline 
	\textbf{\textit{Training and Validation.}} The OCT volumes are divided into a training (7290 scans), validation (1608) and testing (1472) subsets, while it is ensured that eyes belonging to the same patient are not split across subsets. A 20 B-scans from each volume are randomly selected, each has a size of $1024\times200$ and normalized to have pixel values from 0 to 1. A random window around retinal layers with size $84\times84$ is selected with spacing of 4 and 2 in the $y$ and $x$ directions in order. Then, a random rigid transformation is applied on each cropped window separately. The range of simulated transformation parameters is chosen to be from -5 to 5 for $T_x$, $T_y$, and $T_\theta$. This is to cover all possible eye movements. The cropped B-scan and its corresponding transformed B-scan represent the environment that the agent interacts with throughout its lifetime (i.e. one episode). During training, we use a replay experience memory of size $1e6$ to store transitions of $(s_t,a_t,r_t,s_{t+1})$. Then a batch of size 256 is randomly selected. Each input sample has the size $84\times84\times4$, where 4 represents the previous action steps taken by the agent. The network is trained using Adam optimizer with a learning rate of $1e^{-3}$ for $125$ epochs, each has $20,000$ steps with a maximum of $200$ steps per episode. 
	\section{Experimental Results and Discussion}

\begin{figure}
		\centering 
		%\scriptsize
		\setlength\extrarowheight{1pt}
		\setlength\tabcolsep{0pt}
		\begin{tabular}{ccc}
			\includegraphics[width=0.33\linewidth]{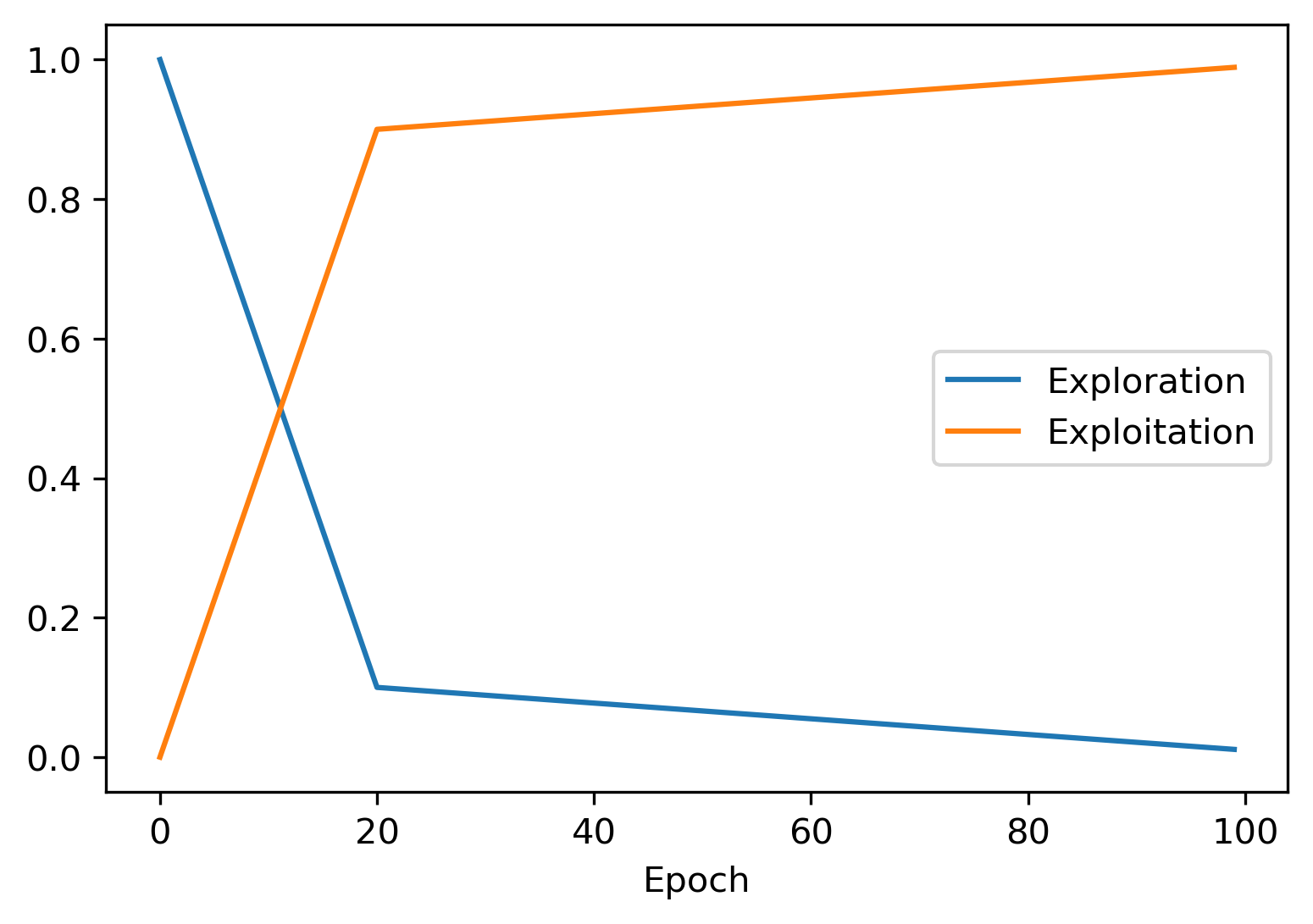} & 
			\includegraphics[width=0.33\linewidth]{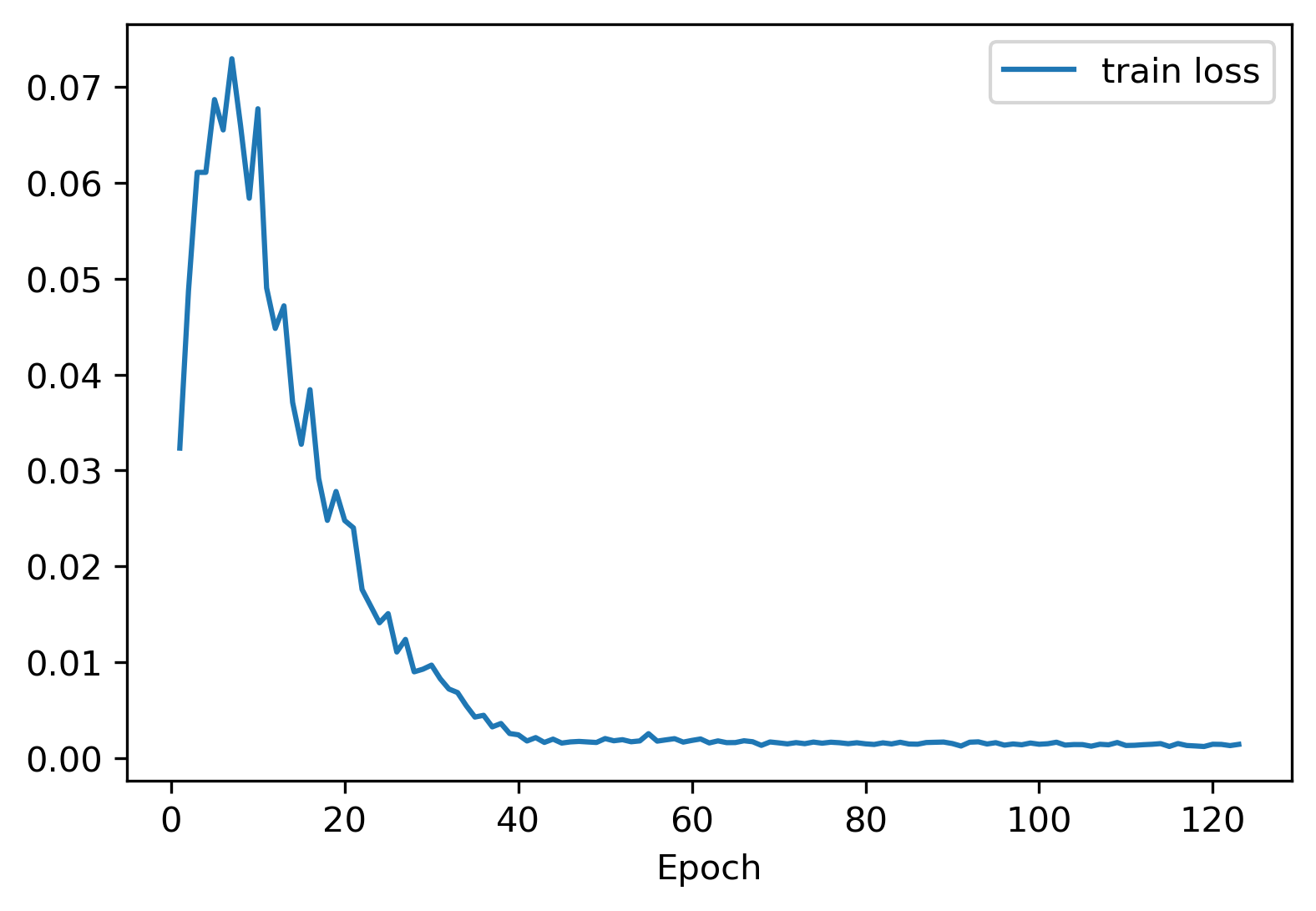} & 
			\includegraphics[width=0.33\linewidth]{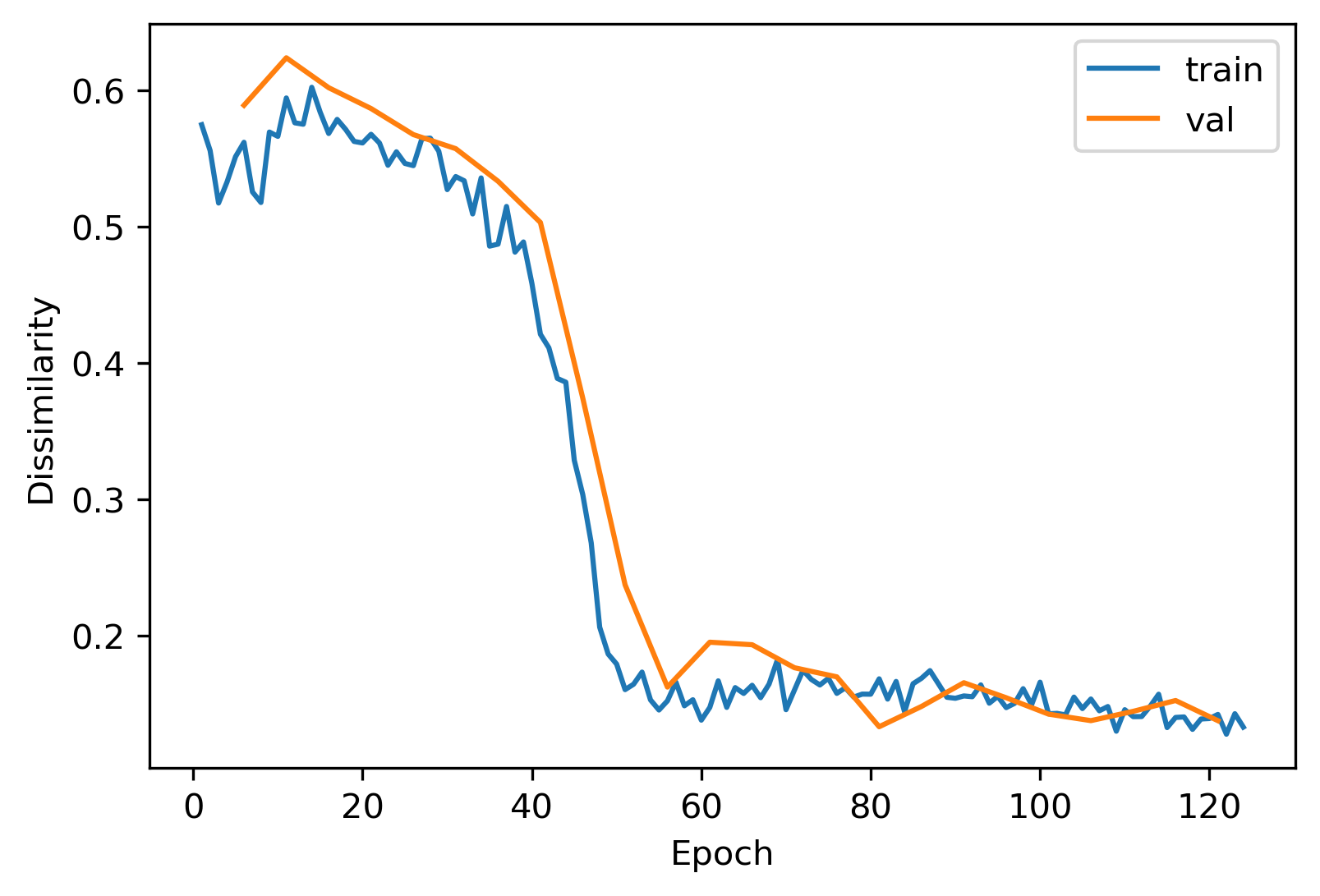} \\
			(a) & (b) & (c)
		\end{tabular}
		%\vspace{-3ex}
		\caption{Artificial agent training graphs. 	(a) Exploration and exploitation rates, (b) Training loss, and (c) Image dissimilarity measures}
		\label{fig:tr_curves}
	\end{figure}

	The proposed DDQN-OCT model is implemented using Python and TensorFlow %\cite{gulli2017deep} 
	on a single V100 GPU. 
	The exploration rate for the artificial agent starts with 1 and linearly decreases to reach 0.1 in epoch \#20, followed by another linear decrease till epoch \#100 as shown in Figure \ref{fig:tr_curves}-(a). Also, Figure \ref{fig:tr_curves}-(b) and (c) plot the training and validation loss curves, and image distances in order, which show a very good convergence for the model without overfitting. 
	%\vspace{-3ex}
	
\begin{figure}[t]
		\centering 
		%\scriptsize
		\setlength\extrarowheight{0pt}
		\setlength\tabcolsep{1pt}
		\begin{tabular}{cc}
			\includegraphics[width=0.5\linewidth,height=1.5cm]{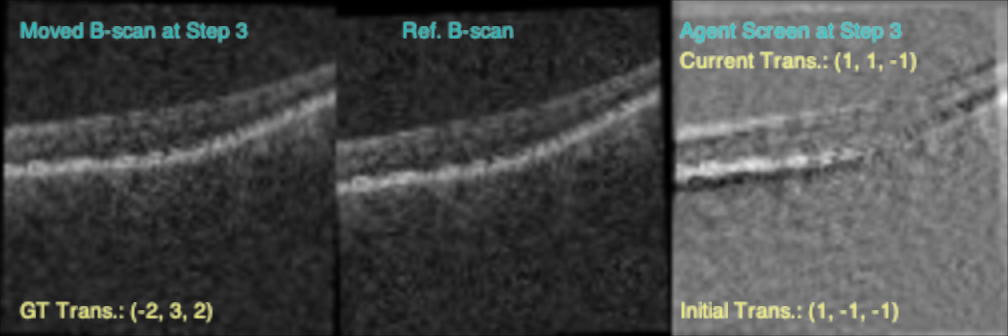} & 
			\includegraphics[width=0.5\linewidth,height=1.5cm]{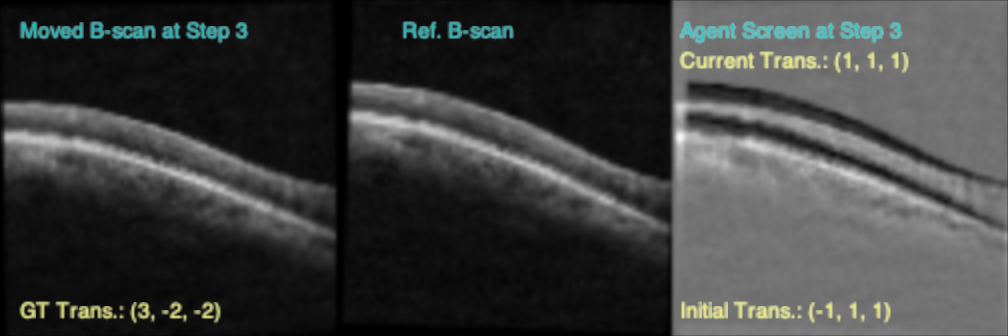} \\
			\includegraphics[width=0.5\linewidth,height=1.5cm]{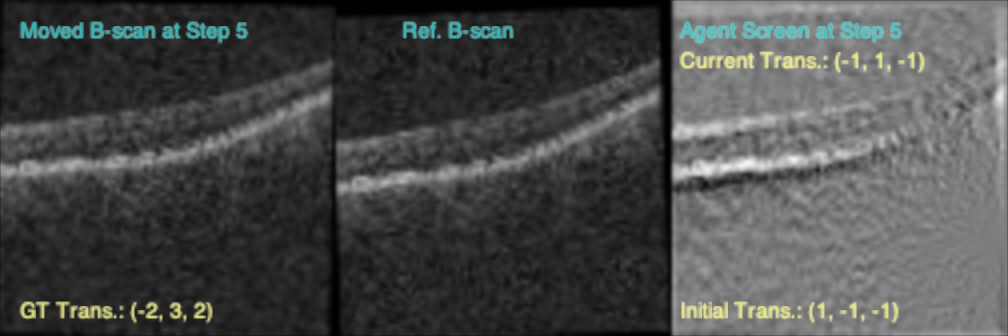} & 
			\includegraphics[width=0.5\linewidth,height=1.5cm]{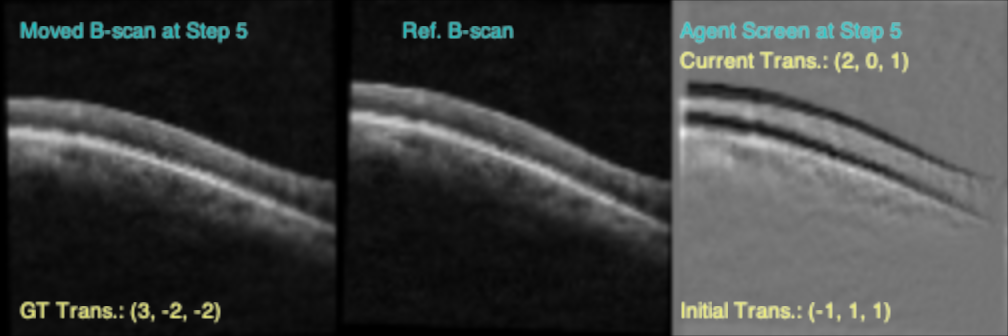} \\
			\includegraphics[width=0.5\linewidth,height=1.5cm]{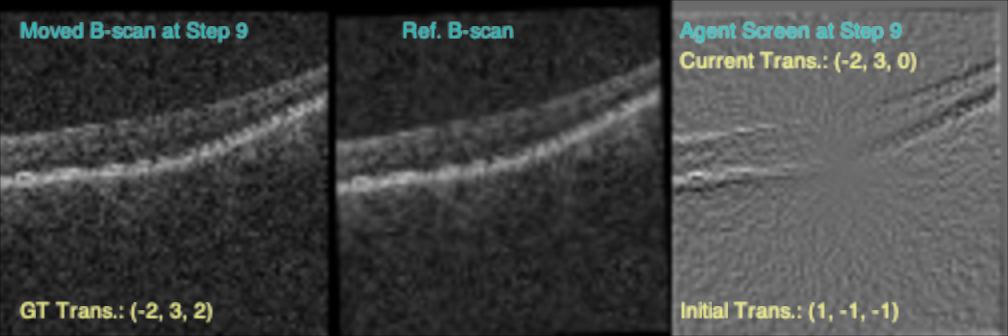} & 
			\includegraphics[width=0.5\linewidth,height=1.5cm]{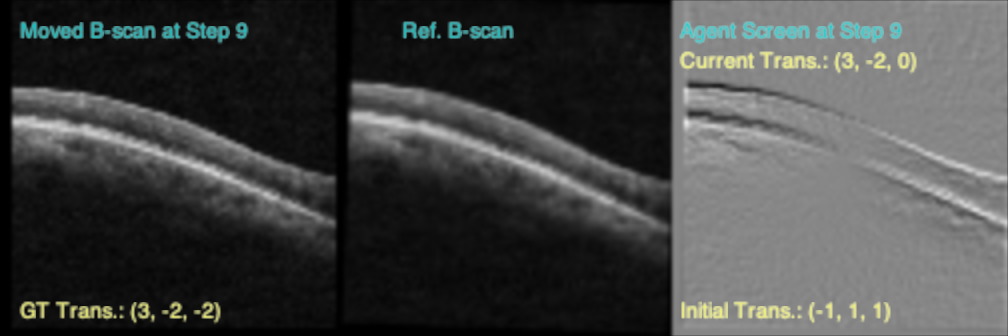} \\
			\includegraphics[width=0.5\linewidth,height=1.5cm]{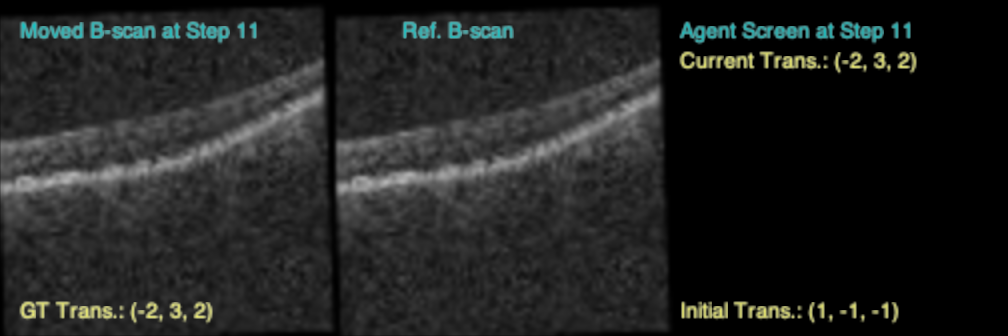} & 
			\includegraphics[width=0.5\linewidth,height=1.5cm]{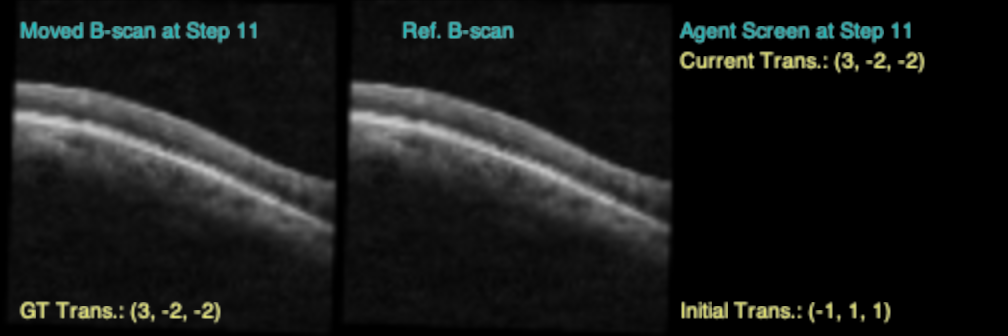} \\   
			(a) Noisy B-scan& (b) Denoised B-scan
		\end{tabular}
		%\vspace{-3ex}
		\caption{Visual results for the proposed DDQN-OCT model}
		\label{fig:visual_proposed}
	\end{figure}

	%\vspace{-3ex}
	To visualize how our method works, we test DDQN-OCT model trained on noisy scans, on both noisy and denoised B-scans. We denoise the B-scans using the Generative Adversarial Network (GAN) model proposed in \cite{halupka2018retinal}. Instead of cropping a random window, we resize the whole B-scan to match the network input shape (See Figure \ref{fig:visual_proposed}). The figure shows the agent results using randomly selected noisy B-scan (left column) and denoised B-scan (right column). The figure has 4 rows, each row displays the reference B-scan, aligned B-scan and agent screen at steps 3, 5, 9, and 11. From the figure, the agent reaches the best alignment after 11 steps. Also, the ground truth transformation parameters and agent current transformation are displayed in yellow. 
	
	For quantitative evaluation, we compute normalized mutual information (NMI), cross-correlation coefficient ($\rho$), agent score (i.e. cumulative reward) and execution time for each sample in our test set that contains 29,440 B-scans. We then compute the average, standard deviation, lower, middle, and upper quartiles for each statistical measure across the test set as reported in Table \ref{tab:proposed_stats}. The proposed model achieves an average of 0.985 and 0.914 for NMI and $\rho$, respectively. Furthermore, to quantify the impact of speckle noise on the agent's training, we retrain the model using denoised B-scans. Statistical results show an increase of 4\% in correlation measure ($\rho$), while a decrease of 2\% for NMI measure (Table \ref{tab:proposed_stats}).  
	
	\begin{table}
		\centering
		%\scriptsize
		\setlength\tabcolsep{5pt}
		\setlength\extrarowheight{1pt}
		\caption{Statistical measures for our proposed DDQN-OCT model }
		%\vspace{-3ex}
		\begin{tabular}{|c|l|cccc|}
			\hline 
			&Statistical measure &        NMI &      $\rho$ & Episode score & Time (sec) \\	\hline
			\multirow{4}{*}{\rotatebox[origin=c]{90}{\parbox[c]{1cm}{\centering Noisy B-scans}}} & Average $\pm$ Std&  0.985 $\pm$ 0.064&  0.914 $\pm$ 0.128&  3.343 $\pm$ 3.036 & 0.445 $\pm$ 0.498 \\
			&Lower quartile (25\%)  &  0.990 &  0.826 &  0.218 & 0.040\\
			&Median (50\%)  &  1.00 &  1.00 &  5.455 & 0.061 \\
			&Upper quartile (75\%) &  1.00 &  1.00  &  5.562& 1.027 \\ \hline
			\multirow{4}{*}{\rotatebox[origin=c]{90}{\parbox[c]{1cm}{\centering Denoised B-scans}}} &Average $\pm$ Std &  0.969 $\pm$ 0.094 &  0.957 $\pm$ 0.076&   3.621 $\pm$  2.623 & 0.607  $\pm$  0.787 \\
			&Lower quartile (25\%)  &  0.986 &  0.945 &    0.280 & 0.058 \\
			&Median (50\%) &  0.989 &  0.984 &   5.254 & 0.105  \\
			&Upper quartile (75\%)  &  1.00 &  1.00 &  5.347 & 1.195\\ \hline
		\end{tabular}
		%\vspace{-3ex}
		\label{tab:proposed_stats}
	\end{table}
	
	As a comparative study, we evaluate the performance of \textit{elastix} intensity-based medical image registration approach, described in \cite{klein2009elastix}. We use same test set and also evaluate using noisy and denoised B-scans separately. Results are reported in Table \ref{tab:affine_stats}, where our DDQN-OCT model has a significant improvement comparing to \textit{elastix} registration approach with more than 50\% and 10\% for NMI and $\rho$, respectively. The table also shows that \textit{elastix} approach works much better on denoised scans than noisy scans with an improvement of 10\% and 4\% for NMI and $\rho$, respectively. We also record the execution time for all our experiments as shown in Tables \ref{tab:proposed_stats} and \ref{tab:affine_stats} where our model takes much less time than \textit{elastix} approach with an average of 0.5 second per images.
	
	\begin{table}
		\centering
		%\scriptsize
		\setlength\tabcolsep{5pt}
		\setlength\extrarowheight{1pt}
		\caption{Statistical measures for the literature approach: \textit{elastix} registration}
		%\vspace{-3ex}
		\begin{tabular}{|c|l|ccc|}
			\hline 
			&Statistical measure &        NMI &      $\rho$ & Time (sec) \\	\hline
			\multirow{4}{*}{\rotatebox[origin=c]{90}{\parbox[c]{1cm}{\centering Noisy B-scans}}}  & Average $\pm$ Std &  0.344 $\pm$ 0.140 &  0.814 $\pm$ 0.083 &  38.322 $\pm$  14.978 \\
			&Lower quartile (25\%)  &   0.281 &  0.762 &  35.545 \\
			&Median (50\%) &  0.306 &  0.826 &  36.581 \\
			&Upper quartile (75\%)  &  0.331 &  0.875 &  37.867 \\ \hline
			\multirow{4}{*}{\rotatebox[origin=c]{90}{\parbox[c]{1cm}{\centering Denoised B-scans}}} & Average $\pm$ Std&  0.448 $\pm$  0.103&  0.847 $\pm$ 0.072 &  6.840 $\pm$ 0.392\\
			&Lower quartile (25\%)  & 0.386 &  0.802 &  6.580 \\
			&Median (50\%)  &  0.421 &  0.854 &  6.851 \\
			&Upper quartile (75\%) & 0.465 &  0.900 &  7.063 \\ \hline
		\end{tabular}
		\label{tab:affine_stats}
	\end{table}

 We also compare the proposed dueling DQN with other DQN architectures that have been recently presented in the literature. For example, Van Hasselt et al. \cite{van2016deep} proposed a double DQN, by decoupling the selected action from the target network that reduced the observed overestimation and better performance. Also, in \cite{alansary2019evaluating}, combination of double dueling approaches have shown to outperform the original DQN. In this experiment, We train four DQN variants namely, DQN, Double DQN, Dueling DQN and Double Dueling DQN. For evaluation, we apply 2d rigid transformation on random crops from our test set that contains 29,440 B-scans (within 10 pixels and 10 degrees for a crop size of 84$\times$84). Performance measures are reported in Table \ref{tab:proposed_stats1} which shows a very slight performance differences between the variants. Double DQN has the best performance with NMI and $\rho$ of 0.98 and 0.97 in order. Also, original DQN has achieved the least performance which aligns with the results presented in \cite{wang2015dueling,van2016deep,alansary2019evaluating}.
 
 Furthermore, we compare our unsupervised rewarding approach with the supervised one as in \cite{liao2017artificial} (i.e. using ground truth transformation parameters). Surprisingly, the unsupervised rewarding approach outperforms the supervised based training with roughly 4\% improvement.
 
 \begin{table}
 	\centering
 	%\scriptsize
 	\setlength\tabcolsep{5pt}
 	\setlength\extrarowheight{2pt}
 	\caption{Statistical measures for different variants of DQN with supervised and unsupervised training using 29,440 B-scans for evaluation}
 	%\vspace{-3ex} % Statistical measure Average $\pm$ Std
 	\begin{tabular}{|c|cccc|}
 		\hline 
 		& NMI 		&      $\rho$ & Episode score & Time (sec) \\	\hline
 		\multicolumn{5}{|c|}{\textbf{\textit{Unsupervised Training}}} \\ \hline
 		DQN & 0.969$\pm$ 0.083 & 0.958 $\pm$ 0.077 &  3.796 $\pm$ 2.687  & 0.335$\pm$0.469 \\
 		Dueling DQN& 0.973$\pm$0.085 &  0.959$\pm$ 0.080&  9.475$\pm$4.341  & 0.249$\pm$ 0.265 \\
 		Double & 0.978$\pm$0.070 &  0.965$\pm$0.066  &  9.738$\pm$4.487 & 0.198$\pm$ 0.216\\
 		Double Dueling DQN & 0.974$\pm$0.081 &  0.960$\pm$0.075 &  9.594$\pm$4.010  & 0.248$\pm$0.264 \\ \hline	
 		\multicolumn{5}{|c|}{\textbf{\textit{Supervised Training}}} \\ \hline
 		Dueling DQN & 0.934$\pm$0.147 &  0.902 $\pm$ 0.121 &  5.979$\pm$13.399 & 0.388$\pm$0.218 \\ \hline
 	\end{tabular}
 	\vspace{-3ex}
 	\label{tab:proposed_stats1}
 \end{table}
 
	\section*{Conclusion}
	Registration is a critical step in automated analysis of medical images for monitoring patients, and access to an accurate unsupervised registration method is of immense value, given the costly practice of curating annotated images as needed in supervised methods. In this paper, we lay out a novel framework for unsupervised 2D rigid registration of medical images, in particular OCT volumes of retina, which takes advantage of intensity-based techniques, resulting to state-of-the-art performance. In doing so, an artificial agent is presented that is able to efficiently align two B-scans by finding the 2D transformation parameters. The agent is trained using dueling deep Q-network in an unsupervised manner, where a combination of intensity based image similarity measures are used to guide the rewarding system. The proposed DDQN-OCT model markedly outperforms the \textit{elastix} intensity based medical image registration approach. Also, the proposed framework has shown the strong potential to be applied to other applications.
	
	\section*{Bibliography}
	
\bibliographystyle{splncs04}
\bibliography{all}
	
\end{document}